\documentclass[twoside]{article}
\makeatletter\if@twocolumn\PassOptionsToPackage{switch}{lineno}\else\fi\makeatother

\usepackage{hyperref}
\usepackage{subfigure}

\usepackage[left]{lineno}

\usepackage{tabulary,graphicx,times,caption,fancyhdr,amsfonts,amssymb,amsbsy,latexsym,amsmath}
\usepackage[utf8]{inputenc}
\usepackage{url,multirow,morefloats,floatflt,cancel,tfrupee}
\makeatletter

\AtBeginDocument{\@ifpackageloaded{textcomp}{}{\usepackage{textcomp}}}
\makeatother
\usepackage{colortbl}
\usepackage{xcolor}
\usepackage{pifont}
\usepackage[nointegrals]{wasysym}
\makeatletter

\def\mcWidth#1{\csname TY@F#1\endcsname+\tabcolsep}

\def\cAlignHack{\rightskip\@flushglue\leftskip\@flushglue\parindent\z@\parfillskip\z@skip}
\def\rAlignHack{\rightskip\z@skip\leftskip\@flushglue \parindent\z@\parfillskip\z@skip}

\@ifundefined{etal}{}{}

\usepackage{ifxetex}
\ifxetex\else\if@twocolumn\@ifpackageloaded{stfloats}{}{\usepackage{dblfloatfix}}\fi\fi

\AtBeginDocument{
\expandafter\ifx\csname eqalign\endcsname\relax
\def\eqalign#1{\null\vcenter{\def\\{\cr}\openup\jot\m@th
  \ialign{\strut$\displaystyle{##}$\hfil&$\displaystyle{{}##}$\hfil
      \crcr#1\crcr}}\,}
\fi
}

\AtBeginDocument{%
  \@ifpackageloaded{endfloat}%
   {\renewcommand\efloat@iwrite[1]{\immediate\expandafter\protected@write\csname efloat@post#1\endcsname{}}}{\newif\ifefloat@tables}%
}%

\def\BreakURLText#1{\@tfor\brk@tempa:=#1\do{\brk@tempa\hskip0pt}}
\let\lt=<
\let\gt=>
\def\processVert{\ifmmode|\else\textbar\fi}

\@ifundefined{subparagraph}{
\def\subparagraph{\@startsection{paragraph}{5}{2\parindent}{0ex plus 0.1ex minus 0.1ex}%
{0ex}{\normalfont\small\itshape}}%
}{}

\newcommand\role[1]{\unskip}
\newcommand\aucollab[1]{\unskip}
  
\@ifundefined{tsGraphicsScaleX}{\gdef\tsGraphicsScaleX{1}}{}
\@ifundefined{tsGraphicsScaleY}{\gdef\tsGraphicsScaleY{.9}}{}
\def\checkGraphicsWidth{\ifdim\Gin@nat@width>\linewidth
	\tsGraphicsScaleX\linewidth\else\Gin@nat@width\fi}

\def\checkGraphicsHeight{\ifdim\Gin@nat@height>.9\textheight
	\tsGraphicsScaleY\textheight\else\Gin@nat@height\fi}

\def\fixFloatSize#1{}
\let\ts@includegraphics\includegraphics

\def\inlinegraphic[#1]#2{{\edef\@tempa{#1}\edef\baseline@shift{\ifx\@tempa\@empty0\else#1\fi}\edef\tempZ{\the\numexpr(\numexpr(\baseline@shift*\f@size/100))}\protect\raisebox{\tempZ pt}{\ts@includegraphics{#2}}}}

\AtBeginDocument{\def\includegraphics{\@ifnextchar[{\ts@includegraphics}{\ts@includegraphics[width=\checkGraphicsWidth,height=\checkGraphicsHeight,keepaspectratio]}}}

\DeclareMathAlphabet{\mathpzc}{OT1}{pzc}{m}{it}

\def\URL#1#2{\@ifundefined{href}{#2}{\href{#1}{#2}}}

\def\UrlOrds{\do\*\do\-\do\~\do\'\do\"\do\-}%
\g@addto@macro{\UrlBreaks}{\UrlOrds}

\edef\fntEncoding{\f@encoding}

\makeatother

\newif\ifmultipleabstract\multipleabstractfalse%
\usepackage[paperheight=11in,paperwidth=8.3in,margin=2.5cm,headsep=.7cm,top=2.5cm]{geometry}
\usepackage[T1]{fontenc}

\renewenvironment{abstract}
{\vspace*{-1pc}\trivlist\item[]\leftskip\hindawiIndent\par\vskip4pt\noindent\textbf{\abstractname}\mbox{\null}\\}{\par\noindent\endtrivlist}

 \date{} \emergencystretch 8pt

\makeatletter\def\hindawiIndent{0pc}
\def\author#1{\gdef\@author{\hskip-\dimexpr(\tabcolsep)\hskip\hindawiIndent\parbox{\dimexpr\textwidth-\hindawiIndent}{\raggedright\bfseries#1}}}

\def\title#1{\gdef\@title{\vspace*{-30pt}\raggedright\textbf{ \journaltitle}~\\\raggedright\bfseries\ifx\@articleType\@empty\vspace*{20pt}\else\vspace*{20pt}\@articleType\vspace*{20pt}\\\fi#1}}
\let\@articleType\@empty \def\articletype#1{\gdef\@articleType{{\normalfont\itshape#1}}}

\let\@runningHead\@empty \def\RunningHead#1{\gdef\@runningHead{{\normalfont #1}}}

\usepackage[numbers,sort&compress]{natbib}

\def\journaltitle{}

\makeatletter
\@ifpackageloaded{geometry}{\geometry{a4paper}}{\usepackage[a4paper]{geometry}}
\makeatother

\begin{document}

\title{A Software to Detect OCC Emotion, Big-Five Personality and Hofstede Cultural Dimensions of Pedestrians from Video Sequences}

\author{Rodolfo Migon Favaretto$^{*,1}$, Victor Araujo$^{1}$, Soraia Raupp Musse$^{1}$, Felipe Vilanova$^{2}$ and Angelo Brandelli Costa$^{2}$}

\maketitle 

\begin{abstract}
This paper presents a video analysis application to detect personality, emotion and cultural aspects from pedestrians in video sequences, along with a visualizer of features. The proposed model considers a series of characteristics of the pedestrians and the crowd, such as number and size of groups, distances, speeds, among others, and performs the mapping of these characteristics in personalities, emotions and cultural aspects, considering the Cultural Dimensions of Hofstede (HCD), the Big-Five Personality Model (OCEAN) and the OCC Emotional Model. The main hypothesis is that there is a relationship between so-called intrinsic human variables (such as emotion) and the way people behave in space and time. The software was tested in a set of videos from different countries and results seem promising in order to identify these three different levels of psychological traits in the filmed sequences. In addition, the data of the people present in the videos can be seen in a crowd viewer.
\end{abstract}
    
\section*{Introduction}

Crowd analysis is a phenomenon of great interest and can benefit areas as Surveillance, entertainment and the social sciences. Currently, with the progress on video processing and computing technology systems, it is possible to develop algorithms to detect and compute pedestrian features in videos. Literature presents many applications of crowd analysis, like counting people in crowds \cite{Chan2009,cai2014}, group and crowd movement and formation~\cite{Solmaz2012,Zhou2014,Ricky:15} and detection of social groups~\cite{solera_2013,Shao2014,Feng2015,Chandran2015}. Normally, these approaches are based on tracking or optical flow algorithms, and handle with features like walking speed, directions and distances over time.

However, there are attributes that can affect such pedestrian features, such as personality and cultural aspects. On this subject, Chattaraj et al.~\cite{Chattaraj:2009} suggested that cultural and population differences could produce deviations in speed, density and flow of the crowd. In~\cite{Favaretto:2019:CAV}, the authors presents a study regarding group behavior in a controlled experiment focused on differences in an important attribute that vary across cultures - the personal spaces, this study was conducted in two countries: Brazil and Germany. Cultural aspects are specific to a group of individuals. Interested in these aspects, Geert Hofstede created a model which became known as the Cultural Dimensions of Hofstede (HCD)~\cite{Hofstede:2001}. HCD is a model with six cultural dimensions that seek to describe the effects of a society's culture on the values of its members and how these values relate to their behaviors~\cite{Hofstede:2011}.

Another attribute that interfere in the cultural aspects of crowd is the personality of its pedestrians. Personality is specific to each individual~\cite{Hofstede:2001}. During the past decades, different personality models have been proposed and studied by psychologists~\cite{Abele:2007, Mccrae:1992, Goldberg:1990}. The model known as Big-Five~\cite{Mccrae:1992}, also referenced as OCEAN is the most adopted and used. OCEAN is a descriptive (taxonomic) psychological model of five personality trait factors that were discovered empirically in 1981~\cite{Goldberg:1982}.

Saifi et al.~\cite{Saifi2016} proposed a mapping between psychological traits and emotions. In their work, the OCEAN~\cite{Mccrae:1992} personality model was combined with the OCC~\cite{occ1990} emotional model in order to find the susceptibility of each of the five OCEAN personality factors to feel each OCC emotion. The OCC (Ortony, Clore, and Collins) emotional model indicate the susceptibility of each of the five personality factors to feeling every emotion.

Regarding game research area, cultural features can determine behavior of simulated crowds~\cite{knobvisualization, dihl2017generating}, character animations~\cite{durupinar2016psychological}, or assist in the study of human perception about simulated crowds~\cite{araujo2019much, Araujo:2019:IVA, yang2018you}. All such cases can be used in serious game contexts to assist psychologists in the analysis of cultural features. In the context of entertainment, these features could be present in games with analysis of scenes and possible different actions of characters (e.g., criminal analysis), as in the game Detroit~\cite{detroit}. In Detroit, there are times when it takes the behavioral analysis of androids characters to make a decision about them. However, if the androids presented cultural features, the analyzes could become more complex and closer to reality.

Based on previously work we develop an application called \emph{GeoMind}\footnote{Download and more information about how to use \emph{GeoMind} can be found at {http://rmfavaretto.pro.br/geomind}.} which aims to automatically detect personality, emotion and cultural aspects from pedestrians in crowd videos. Next section (Section~\refname{sec:related}) discusses some related work, in Section~\ref{sec:prop_approach} we present details about the proposed approach and in Section~\ref{sec:soft_concept} we show how \emph{GeoMind} works. Final considerations and future work are addressed in Section~\ref{sec:final_considerations}.

\section{Related work}
\label{sec:related}

The cultural influence can be represent in crowds attributes as adopted personal spaces, preferred speed, the way pedestrians avoid collisions, group formations~\cite{Fridman2013} and others. Many work~\cite{Weina:2012, Solmaz2012, Chandran2015} in the literature focus on the identification of groups using computer vision. Ge, Collins and Ruback~\cite{Weina:2012} propose methods to detect small groups of individuals who are walking together. In the work proposed by Chandran et al.~\cite{Chandran2015} a non-recursive motion similarity clustering algorithm is proposed to identify pedestrians traveling together in social groups. These work aim mainly to detect social groups from videos of crowds. Others work, in addition to detect the groups, analyze their behaviors in the space~\cite{Feng2015}.

In our previous work~\cite{Favaretto:2016:Hofstede} we proposed a method to identify groups and characterize them to assess the aspects of cultural differences through the mapping of the Hofstede's dimensions~\cite{Hofstede:2011}. A similar idea, however using computer simulation and not focused on computer vision, is proposed by Lala et al.~\cite{Lala2012}. They use Hofstede's dimensions to create a simulated crowd from a cultural perspective.  In addition to the cultural aspects, people can also be different because they have different personalities. So, some authors are using OCEAN model OCEAN~\cite{Mccrae:1992}.  Gorbova and collaborators~\cite{Gorbova2017}, present a system of automatic personality screening from video presentations in order to make a decision whether a person has to be invited to a job interview based on visual, audio and lexical cues. As we presented in \cite{Favaretto:2017:BigFive}, we proposed a model to detect personality aspects based on the Big-five personality model using individuals behaviors automatically detected in video sequences.

Several models have been developed to explain and quantify basic emotions in humans. One of the most cited is proposed by Paul Ekman~\cite{ekman1971constants} which considers the existence of 6 universal emotions based on cross-cultural facial expressions (anger, disgust, fear, happiness, sadness and surprise). Other approaches such as Affective Neuroscience postulate, from an evolutionary perspective, consider other groups of emotions such as fear, rage/anger and sadness/panic~\cite{montag2017primary}. In~\cite{Favaretto:2018:Emotion} we proposed a way to detect pedestrian emotions in videos, based on OCC emotion model. To detect the emotions of each pedestrian, we used OCEAN as inputs, as proposed by Saifi~\cite{Saifi2016}. The contribution of this work is to present and test a new video analysis application to detect cultural aspects, personalities and emotions of people in video sequences.

\section{The proposed approach}
\label{sec:prop_approach}

Our approach presents four main modules: \textit{i}) pedestrian features extraction, \textit{ii}) OCEAN personality detection, \textit{iii}) OCC emotion detection and \textit{iv}) Hofstede Cultural Dimensions mapping. Each one of these modules is described next.

\subsection{Pedestrian features extraction}

Initially, the information about people must be obtained using a tracker to recover pedestrians trajectories (details about the tracking format used as input in the software are described in Sec.~\ref{sec:soft_concept}). Based on the tracking input file, we compute information for each pedestrian $i$ at each timestep: \textit{i)} 2D position $x_i$ (meters); \textit{ii)} speed $s_i$ (meters/frame); \textit{iii)} angular variation $\alpha_i$ (degrees) w.r.t. a reference vector $\vec{r}=(1,0)$; \textit{iv)} isolation level $\varphi_i$; \textit{v)} socialization level $\vartheta_i$; and \textit{vi)} collectivity $\phi_i$. To compute the collectivity affected in individual $i$ from all $n$ individuals, we computed:

\begin{equation}
    \phi_i = \sum_{j=0}^{n-1} \gamma e^{(-\beta \varpi(i,j)^{2})},
\end{equation}

\noindent and the collectivity between two individuals is calculated as a decay function of $\varpi(i,j) = s(s_i,s_j).w_1+o(\alpha_i,\alpha_j).w_2$, considering $s$ and $o$ respectively the speed and orientation differences between two people $i$ and $j$, and $w_1$ and $w_2$ are constants that should regulate the offset in meters and radians. We have used $w_1=1$ and $w_2=1$. So, values for $\varpi(i,j)$ are included in interval $0\leq \varpi(i,j) \leq 4.34$. $\gamma = 1$ is the maximum collectivity value when $\varpi(i,j)=0$, and $\beta = 0.3$ is empirically defined as decay constant. Hence, $\phi_{i}$ is a value in the interval $[0;1]$.

To compute the socialization level $\vartheta$ we use an artificial neural network (ANN) with a Scaled Conjugate Gradient (SCG) algorithm in the training process to calculate the socialization $\vartheta_i$ level for each individual $i$. The ANN has 3 inputs (collectivity $\phi_i$ of person $i$, mean Euclidean distance from a person $i$ to others $\bar{d_{i,j}}$ and the number of people in the Social Space\footnote{Social space is related to $3.6$ meters~\cite{hall98}.} according to Hall's proxemics~\cite{hall98} around the person $n_i$). In addition, the network has 10 hidden layers and 2 outputs (the probabilities of socialization and non socialization). The final accuracy was 96\%. We used 16.000 samples (70\% of training and 30\% of validating). Once we get the socialization level $\vartheta_i$, we compute the isolation level $\varphi_i = 1 - \vartheta_i$, that corresponds to its inverse. For more details about how this features are obtained, please refer to~\cite{Favaretto:2017:BigFive, Favaretto:2016:SIB}.

For each individual $i$ in a video, we computed the average for all frames and generate a vector $\vec{V_i}$ of extracted data where $\vec{V_i} = \left [x_i, s_i, \alpha_i, \varphi_i, \vartheta_i, \phi_i \right ]$. In the next section we describe how these features are mapped into personality dimensions.

\subsection{Personality}

The five dimensions of OCEAN are: Openness (“the active seeking and appreciation of new experiences”); Conscientiousness (“degree of organization, persistence, control and motivation in goal directed behavior”); Extraversion (“quantity and intensity of energy directed outwards in the social world”); Agreeableness (“the kinds of interaction an individual prefers from compassion to tough mindedness”); and Neuroticism (how much prone to psychological distress the individual is). To detect the OCEAN of each pedestrian, we used the NEO PI-R~\cite{Costa:1992} that is the standard questionnaire measure of the Five Factor Model. We firstly selected NEO PI-R items related to individual-level crowd characteristics and the corresponding OCEAN-factor. For example: “Like being part of crowd at sporting events” corresponding to the factor “Extroversion”.

As we describe in details in~\cite{Favaretto:2017:BigFive}, we proposed a series of empirically defined equations to map pedestrian features to OCEAN dimensions. Firstly, we selected 25 from the 240 items from NEO PI-R inventory that had a direct relationship with crowd behavior. In order to answer the items with data coming from real video sequences, we propose equations that could represent each one of the 25 items with features extracted from videos. For example, in order to represent the item ``1 - Have clear goals, work to them in orderly way'', we consider that the individual $i$ should have a high velocity $s$ and low angular variation $\alpha$ to have answer compatible with 5. So the equation for this item is $Q_1 = s_i+\frac{1}{\alpha_i}$. In this way, we empirically defined equations for all 25 items, as presented in~\cite{Favaretto:2017:BigFive}. In next section we show how we map the OCEAN dimensions into emotions.

\subsection{Emotion}

As we presented in~\cite{Favaretto:2018:Emotion}, we proposed a way to map the OCEAN dimensions of each pedestrian in OCC Emotion model. This mapping is described in Table~\ref{tab:emotionMapping}. In Table~\ref{tab:emotionMapping}, the plus/minus signals along each factor represent the positive/negative value of each one. For example concerning Openness, O+ stands for positive values (i.e. O $\geq$ 0.5) and O- stands for negative values (i.e. O $<$ 0.5)). A positive value for a given factor (i.e. 1) means the stronger the OCEAN trait is, the stronger is the emotion too. A negative value (i.e. -1) does the opposite, therefore, the stronger the factor's value, the weaker is a given emotion. A zero value means that a given emotion is not affected at all by the given factor. To better illustrate, a hypothetical example is given: if an individual has a high value for Extraversion (for example, E = 0.9), following the mapping in Table~\ref{tab:emotionMapping}, this individual can present signals of happiness (i.e. If E+ then Happiness= 1) and should not be angry (i.e. If E+ then Anger= -1).

\begin{table}[htb]
   \renewcommand{\arraystretch}{1}
   \centering
   \caption{Emotion mapping from OCEAN to OCC.}
     \begin{tabular}{ccccc}
     \hline\noalign{\smallskip}
     \textbf{Factor} & \textbf{Fear} & \textbf{Happiness} & \textbf{Sadness} & \textbf{Anger} \\
     \noalign{\smallskip}
        \hline
     O+ & 0 & 0 & 0 & -1 \\
     O- & 0 & 0 & 0 & 1 \\
     C+ & -1 & 0 & 0 & 0 \\ 
     C- & 1 & 0 & 0 & 0 \\ 
     E+ & -1 & 1 & -1 & -1 \\ 
     E- & 1 & 0 & 0 & 0 \\ 
     A+ & 0 & 0 & 0 & -1 \\ 
     A- & 0 & 0 & 0 & 1 \\ 
     N+ & 1 & -1 & 1 & 1 \\ 
     N- & -1 & 1 & -1 & -1 \\ 
        \hline
     \end{tabular}
   \label{tab:emotionMapping}
\end{table}

\subsection{Cultural Aspects}

In order to map pedestrians features in Cultural dimensions, we proposed an approach based on groups characteristics~\cite{Favaretto:2016:Hofstede}. Indeed, collectivism (COL) is a \% of people grouped, while the individualism (IDV) is a \% of lonely people. Regarding PDI, our hypothesis is that individuals that keep close to each other recognize less the group hierarchy, while higher distances between agents can represent a more explicitly hierarchy recognition. Hence, we used the mean group distance to describe these cultural dimension ($\bar{d}_g$). In terms of LTO/STO, the underlying idea is persistence (long-term) as opposed to quick results (short-term). So, we adapted the group orientation to this dimension, meaning that groups with higher values of angular variation result in short-term orientation ($STO=100-LTO$), which are computed as shown in Equation~\ref{eq:LTO}.

\begin{equation}
LTO = \left\{
\begin{array}{ll}
O_k, & \text{~if~} O_k >= 50 \\
100-O_k, & \text{~otherwise}
\end{array}.
\label{eq:LTO}
\right .
\end{equation}

Considering the MAS dimension, we regard that the group cohesion can represent ``a preference for cooperation''. So, higher levels of cohesion represent more femininity values in such dimension. Indeed, we used also LTO to weight the MAS aspect: $MAS = \sigma_1 GC_k + (1-\sigma_1)LTO$,where $\sigma_1=0.5$ is the empirically chosen weight. Finally, the Indulgence vs. restraint dimension has been characterized by the groups speed and collectivism, given by $IND = \rho_1 S_k + (1-\rho_1) COL$, where $\rho_1 = 0.5$ is an empirically chosen weight. Next section describes the developed application.

\section{The software concept}
\label{sec:soft_concept}

\begin{figure}[!ht]
\centering
\includegraphics[width=0.9\linewidth]{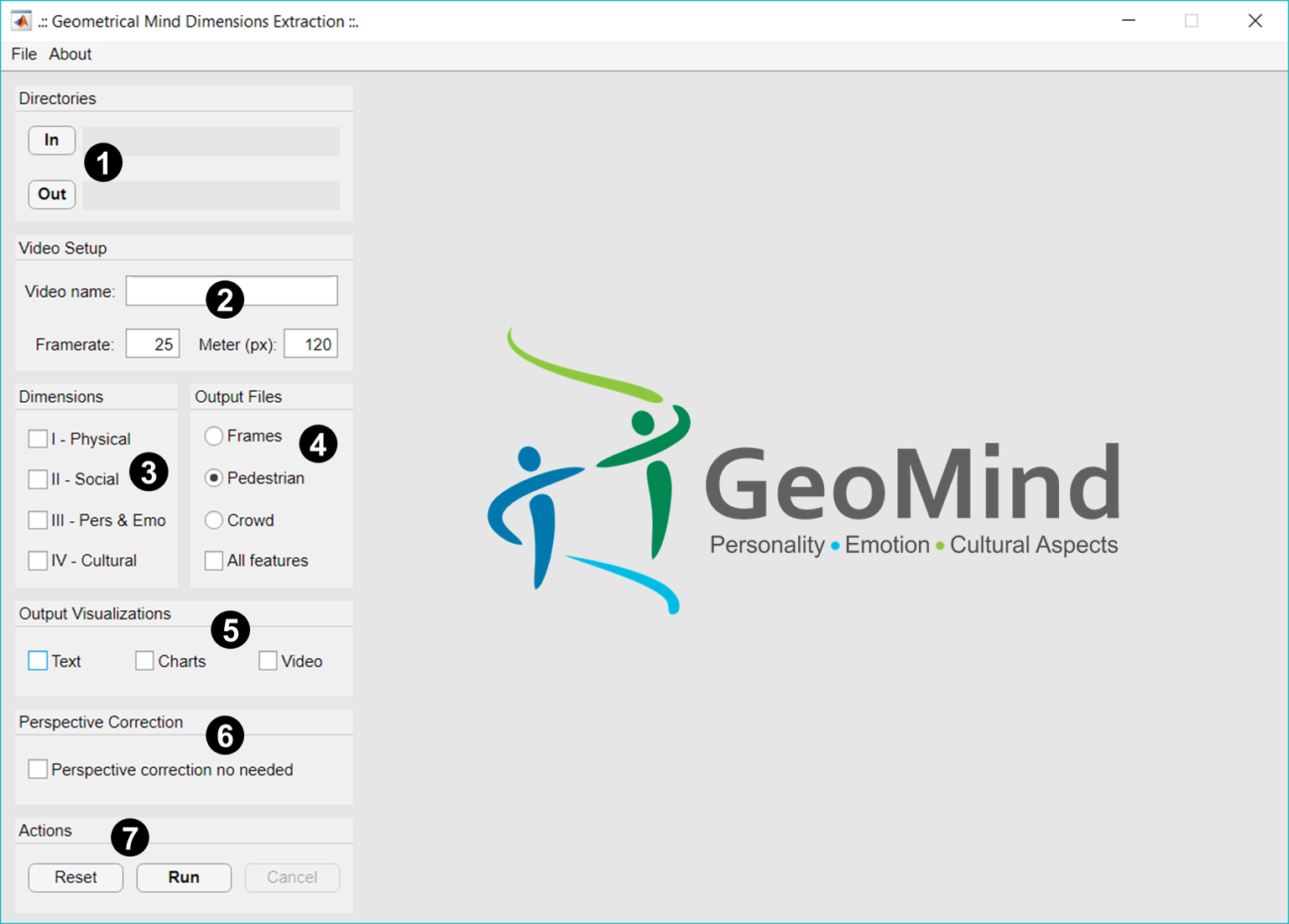}
\caption{Main window of \emph{GeoMind}.}
\label{fig_GeoMind_main}
\end{figure}

The software \emph{GeoMind} was developed using Matlab App Desinger. It was design to be simple and easy to use, allowing users, with a few steps, to obtain a series of pedestrian features from video sequences, based on tracking. Fig.~\ref{fig_GeoMind_main} shows the main window of the software. It is possible to see in the left side the setup panel.

The setup panel (left side of window in Fig.~\ref{fig_GeoMind_main}) contains the input and output configurations of the video processing:

\begin{itemize}
    \item[] \textbf{1. Directories}: The input directory must have the video frames (pictures) and the tracking file. The output files generated by \emph{GeoMind} will be stored in the output directory;
    
    \item[] \textbf{2. Video Setup}: Users must specify a video name, its framerate and how many pixels represent one meter in the video;
    
    \item[] \textbf{3. Dimensions}: The features were grouped into four dimensions, accordingly to its nature: \textit{I - Physical} with geometrical features, such as speed and angular variation; \textit{II - Social} with features related to groups and social behaviours, as collectivity and socialization; \textit{III - Personal and Emotional} with regard to OCEAN personality and OCC emotion models; and \textit{IV - Cultural} with respect to Hofstede cultural Dimensions. The users have to select at least one dimension to be computed and save in the output directory;
    
    \item[] \textbf{4. Output Files}: In this section, users have to select the frequency in which the features are saved in the output files. In addition, they can check the option ``All features", outputting a file with all available information about pedestrians;
    
    \item[] \textbf{5. Output Visualizations}: Users can choose how they want to generate the information about pedestrians, as text in \emph{.txt} files, as plot in charts or in a video;
    
    \item[] \textbf{6. Perspective Correction}: Optionally, users can use a version of the tracking with perspective correction, reducing errors during calculations of pedestrian positions in the video. If this correction is not necessary, just check the ``Perspective correction no needed" option;

    \item[] \textbf{7. Action Buttons}: Once the user filled all the fields in the setup panel, he has 3 options to choose: \textit{Run} to start the video processing; \textit{Reset} to restore default values for each predefined form field; and \textit{Cancel}, to stop the processing in progress.
\end{itemize}

\subsection{Input files format}

The input directory must have the video frames and the tracking file. Each frame of the video sequence must be extracted as an \emph{.jpg} picture. The frame names must have a sequence of six digits, starting by \emph{000000.jpg} (for example, the frame names from a video with 853 frames should be \emph{000000.jpg}, \ldots, \emph{000852.jpg}). The tracking file (named as \emph{tracking.txt}) must have the 2D positions of each pedestrian at each frame.

The tracking file keeps the positions of each pedestrian, in the following format: pedestrian identification P-$<$id$>$, where $<$id$>$ is a sequential number starting in zero; followed by the tuple $<$F X Y$>$, indicating the frame $F$ and the position $(X,Y)$ of that pedestrian in that frame. An example of the tracking file can be found at \emph{GeoMind}'s website. In addition, a tracking file can be used with perspective correction. In this case, the \emph{tracking\_correction.txt} file must have the same format as \emph{tracking.txt}.

\begin{figure}[!ht]
\centering
\subfigure[video][Input files]{\includegraphics[width=0.69\linewidth]{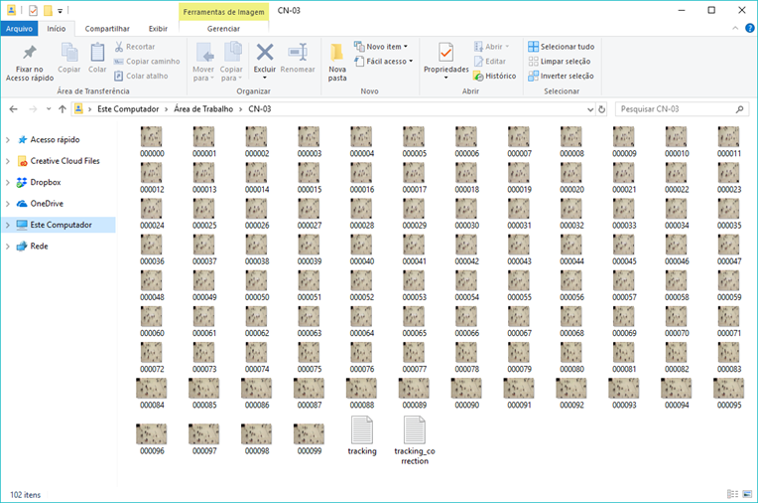}}
\subfigure[video][Tracking file]{\includegraphics[width=0.285\linewidth]{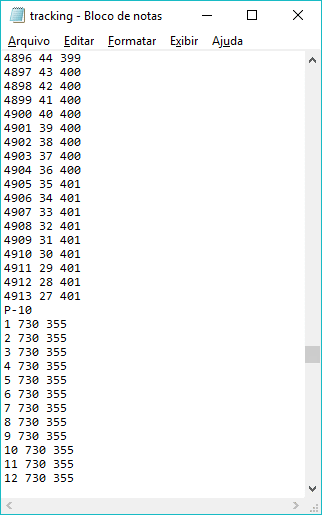}}
\caption{Examples of \emph{GeoMind} input files: input directory with the tracking files and frames of the video (a) and an example of \emph{tracking.txt} file (b).}
\label{fig:files_in}
\end{figure}

Figure~\ref{fig:files_in} illustrates the input files needed by \emph{GeoMind}. Figure~\ref{fig:files_in}(a) shows the input directory, where it is possible to see 100 frames of the video (files \emph{000000.jpg}, \ldots, \emph{000099.jpg}) and the tracking files (\emph{tracking.txt}, with the positions in image coordinates and \emph{tracking\_correction.txt}, with positions in world coordinates, after a perspective correction). Figure~\ref{fig:files_in}(b) shows an example of a tracking file. It is possible to see the last positions of pedestrian \textit{P-9} and the first positions of pedestrian \textit{P-10}.

In the next section we present the results and the files generated by \emph{GeoMind} after the video processing.

\subsection{Software results and file outputs}

\begin{figure}[!ht]
\centering
\includegraphics[width=0.9\linewidth]{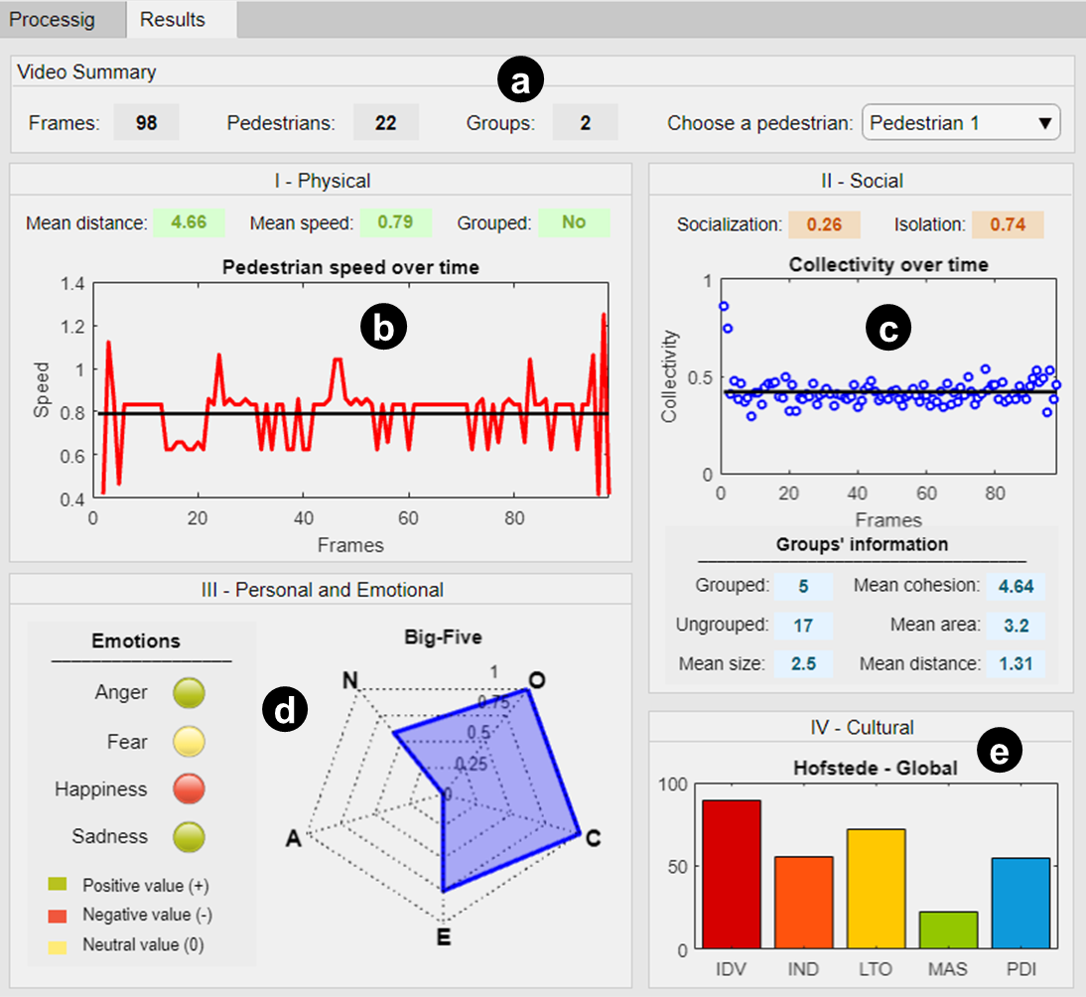}
\caption{Video summary after video processing in \emph{GeoMind}.}
\label{fig_results}
\end{figure}

Fig.~\ref{fig_results} shows a video summary after the software processing. This summary is divided in five areas. Area \textbf{a} shows the video information, such as number of frames, number of pedestrians and number of groups. In this area any individual can be select to see its features (``Pedestrian 1" is selected).

Area \textbf{b} shows a summary on the \textit{Physical} features, presenting the mean distance of selected pedestrian to the others, mean speed, an indication if the pedestrian is part of a group or not and a plot of its speed over time. In area \textbf{c} thee are the features related to the \textit{Social} dimension. It shows data about Isolation and socialization levels and a plot on the collectivity of the selected pedestrian over time. This section also brings information about groups found in the video, such as: number of grouped and ungrouped pedestrians, mean size of the groups (number of pedestrians), mean cohesion, mean area and mean distance between the pedestrians in the group.

\begin{figure}[!ht]
\centering
\subfigure[video][Pedestrians' ID]{\includegraphics[width=0.44\linewidth]{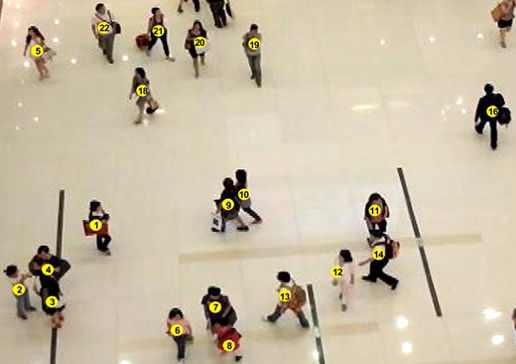}}
\subfigure[video][Plots and text files]{\includegraphics[width=0.44\linewidth]{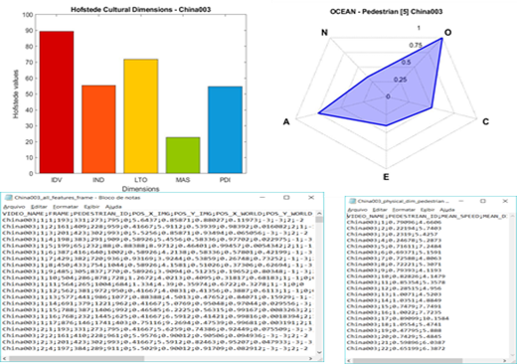}}
\caption{Examples of files generated by \emph{GeoMind}: video with pedestrians ID (a) and plots and text files with features (b).}
\label{fig_files_out}
\end{figure}

Area \textbf{d} shows the \textit{Personal and Emotion} dimensions features, plotting the emotion values and personalities (Big-Five) of the selected pedestrian in the interval $[0, 1]$. Finally, area \textbf{e} is responsible for the \textit{Cultural} features, plotting the Hofstede Cultural Dimensions of the video. Fig.~\ref{fig_files_out} shows some examples of files generated by \emph{GeoMind}: a frame of a video with the pedestrians' ID in Fig.~\ref{fig_files_out}(a) and examples of charts and text files with pedestrians' features in Fig.~\ref{fig_files_out}(b).

One of the files generated by \emph{GeoMind} (\emph{<<video>>\_all\_features\_frame.txt}, where <<video>> indicate the name given to the video in the setup panel at proccessing time) can be used as an input in a visualizer tool. That visualizer tool is presented in next section.

\section{Visualizing Cultural Aspects\label{sec:visualizer}}

As mentioned before, one of the outputs generated by the \emph{GeoMind} software is the \emph{all\_features\_frame.txt} file from a specific video. This file contains all the information about each of the pedestrians present in the video at each frame. This file serves as input to a viewer, which allows, in a simulation environment, a way of visualizing a series of features, such as: cultural aspects, personality traits, social aspects, emotions, socialization, isolation, collectivity, among other characteristics.

\begin{figure}[!htb]
    \centering
    \includegraphics[width=0.85\linewidth]{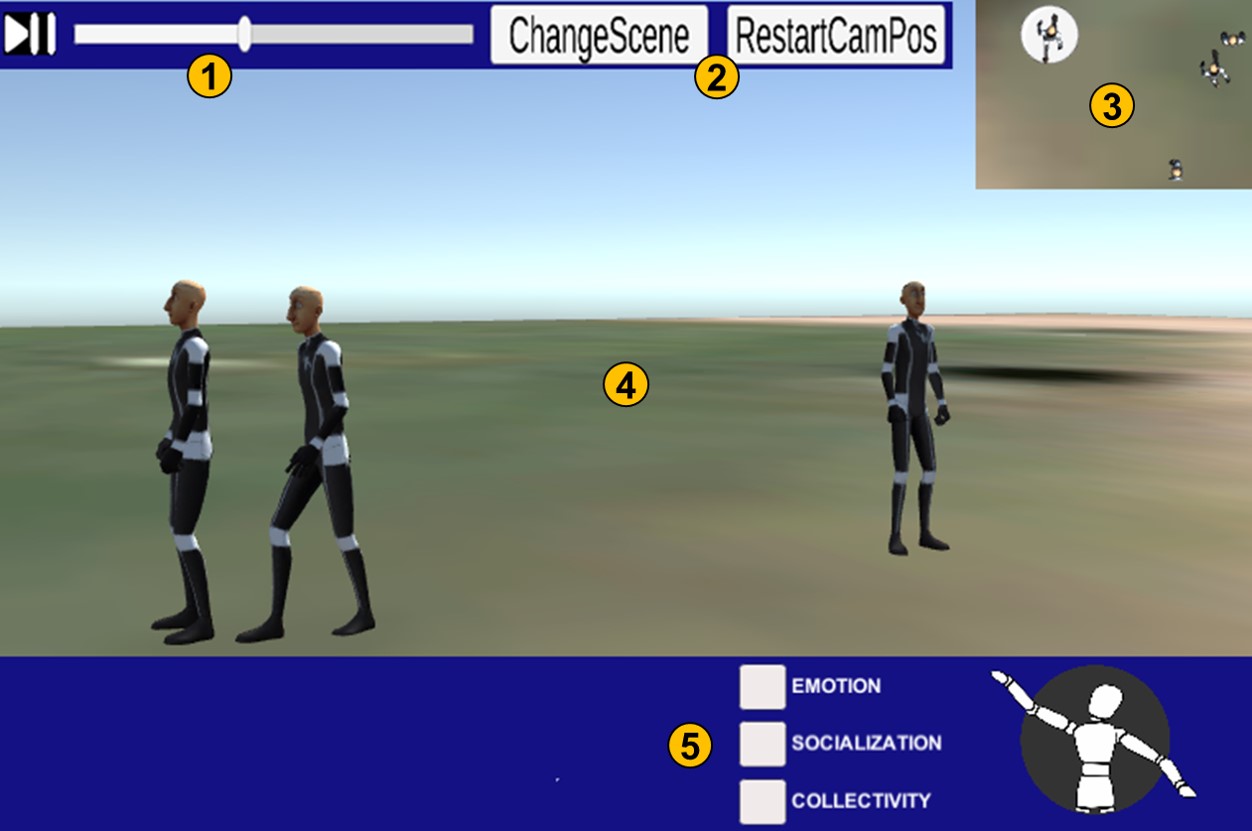}
    \caption{\textbf{Main window of the viewer}: visualization of a file output in \emph{GeoMind} from a determined video.}
    \label{fig:viewer_cam_1p}
\end{figure}

This application was developed using the Unity3D\footnote{Unity3D is available at \url{https://unity3d.com/}} engine, with $C\#$ programming language. The viewer allows the users to rewind, accelerate and stop the simulated video through a time controller, so that the user can observe something that he finds interesting several times, at any time. Figure~\ref{fig:viewer_cam_1p} shows the main window of the viewer.

This environment has three modes of visualization: (i) first-person visualization, where the user can visualize what is happening in the environment from the perspective of a pedestrian present in the video; (ii) top view, where it is possible to observe the movement of the pedestrians through a perspective from above of the video environment; and (iii) an oblique view, where the user can see the entire environment. The viewer is divided in five parts, as follows:
\begin{itemize}
    \item[] \textbf{1. Time controller}:  in the area 1, it is possible to see the button with the start, stop and continue simulation playback functions, together with the frame control bar;
    
    \item[] \textbf{2. Scene setup}: in area 2 there are the \textit{ChangeScene} and \textit{RestartCamPos} buttons, respectively, to go to the viewer's home screen (where the user can load the data file of another video) and restart the camera position for viewing in first person, if it is fixed to a agent, to an initial default position;
    
    \item[] \textbf{3. Top-view cam}: area 3 shows the top view of the environment, as if the user were looking at the crowd through the top;
    
    \item[] \textbf{4. First-person cam}: area 4 shows first-person view from the viewpoint of a previously selected agent. This agent is highlighted in area 3;
    
    \item[] \textbf{5. Features panel}: area 5 is responsible for the features panel, where the users can see up to four selected agents and their features. In addition, it is possible to activate the visualization of the data related to the emotion, socialization and collectivity of agents.
\end{itemize}

\begin{figure}[!htb]
    \centering
    \subfigure[icon][Walking]{\includegraphics[width=0.16\linewidth]{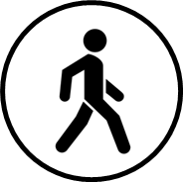}} \hfill
    \subfigure[icon][Running]{\includegraphics[width=0.16\linewidth]{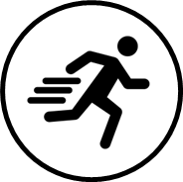}} \hfill
    \subfigure[icon][Socialization]{\includegraphics[width=0.16\linewidth]{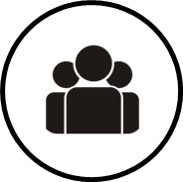}} \hfill
    \subfigure[icon][Isolation]{\includegraphics[width=0.16\linewidth]{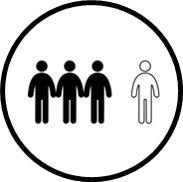}} \hfill
    \subfigure[icon][Individualism]{\includegraphics[width=0.16\linewidth]{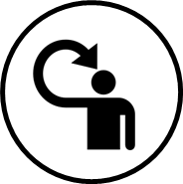}} \\
    
    \subfigure[icon][Collectivism]{\includegraphics[width=0.16\linewidth]{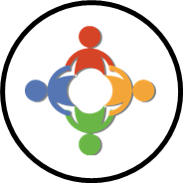}} \hfill
    \subfigure[icon][Anger]{\includegraphics[width=0.16\linewidth]{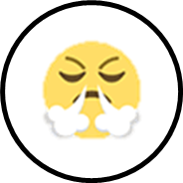}} \hfill
    \subfigure[icon][Fear]{\includegraphics[width=0.16\linewidth]{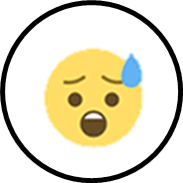}} \hfill
    \subfigure[icon][Happiness]{\includegraphics[width=0.16\linewidth]{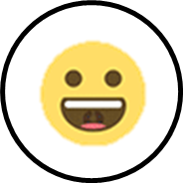}} \hfill
    \subfigure[icon][Sadness]{\includegraphics[width=0.16\linewidth]{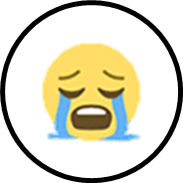}}
    \caption{\textbf{Icons from the features of the viewer}: possible icons shown in the features panel from a determined agent. In (a) and (b) are illustrated the icons that represent the speed of the agent in each frame, respectively, walking and running. In (c) and (d) are illustrated the icons that represent whether the agent is sociable or isolated in that frame. In (e) and (f) are illustrated the icons that represent if the agent is collectivist or individualistic. From (g) to (f) are illustrated the icons that indicate the emotions (anger, fear, happiness, and sadness) of the agent in the current frame.}
    \label{fig:viewer_icons}
\end{figure}

The visualization of agent characteristics is shown in the features panel, illustrated in Figure~\ref{fig:viewer_cam_1p}, area 5. This panel is hidden, only visible on the screen if the mouse cursor passes through the lower region of the screen. In this panel, there are three check-boxes: (i) emotion, (ii) socialization, and (iii) collectivity. The function of these boxes is to enable and disable the visualization of these characteristics in the agents. Figure~\ref{fig:viewer_icons} shows all possible icons that are related to the three options.

This view consists of a set of icons that are displayed at the top of the agents in the simulation. For example, by the time the user selects the emotion field, icons representing the emotions (anger, fear, happiness and sadness) of each agent are displayed on the top of each agent. This icons are displayed in Figure~\ref{fig:viewer_icons}(g-j). 

\begin{figure}[!htb]
    \centering
    \includegraphics[width=0.9\linewidth]{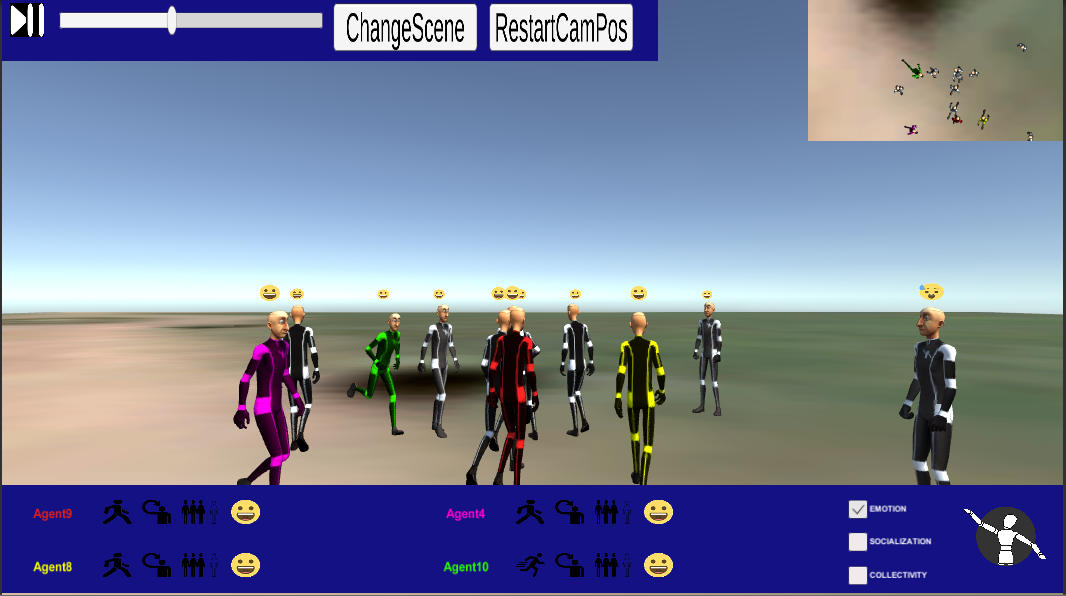}
    \caption{\textbf{Emotion analysis in the viewer}: an example of the emotions shown in the top of each agent. In addition, four agents were select (highlighted with different colors), where is possible to see its features in the panel.}
    \label{fig:viewer_selected}
\end{figure}

Figure~\ref{fig:viewer_selected} shows an example of a video loaded in the viewer. The viewer allows the selection of up to four agents, which are present in the current frame, by right clicking on the humanoids that the user wishes to select. For each selected agent, the color of his clothes is changed and his information is fixed in the features panel, represented by the same color of the clothes and by an identifier (for example \textit{Agent10}, who is highlighted in green).

Besides the agent identifier are the representative icons of their characteristics: speed, whether the agent is walking or running in the current frame; collectivity, whether the agent is collective or not; socialization, whether the agent is sociable or isolated; and emotion, whether the agent is angry, happy, sad, or afraid. As an example, \textit{Agent10} (highlighted in green in Figure~\ref{fig:viewer_selected}) which is running, is not a collective agent, is isolated and happy. All possible icons are presented in Figure~\ref{fig:viewer_icons}.

\begin{figure}[!htb]
    \centering
    \includegraphics[width=0.9\linewidth]{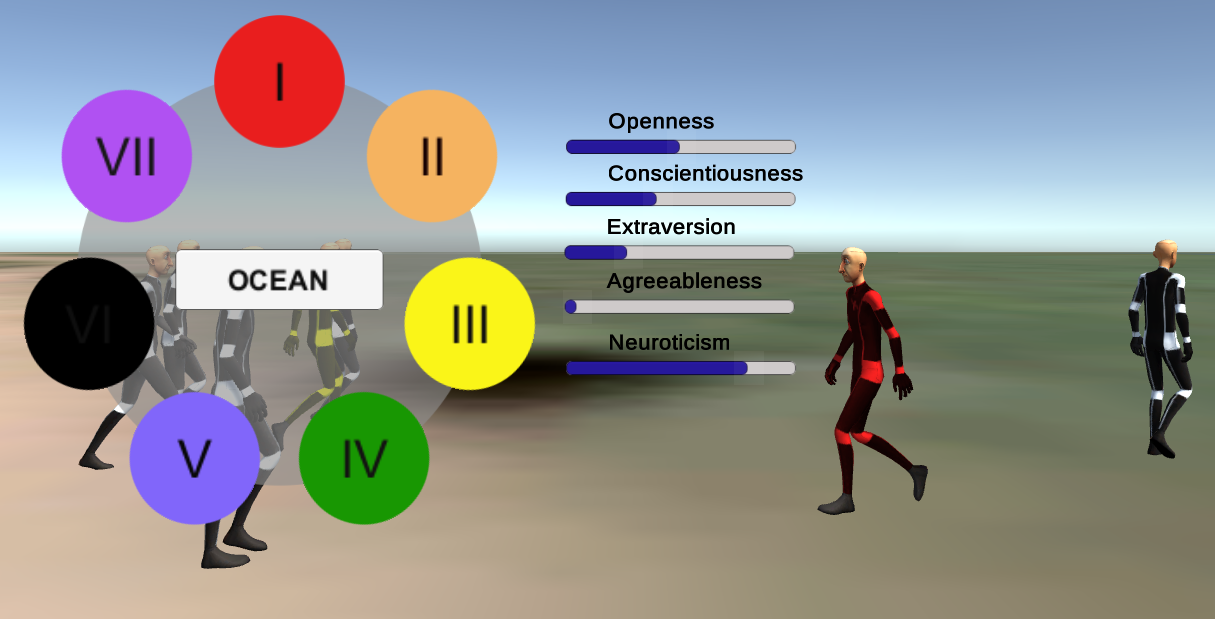}
    \caption{\textbf{Radial menu of features}: an example of the personalities shown in the radial menu from a selected pedestrian.}
    \label{fig:viewer_radial_menu}
\end{figure}

Besides that, the viewer also provides a radial menu to show the features' values of a selected pedestrian. For this, it is enough that the user clicks on the identifier of a certain agent in the panel of features and the radial menu will appear. Figure~\ref{fig:viewer_radial_menu} shows an example of the menu. 

The features illustrated in Figure~\ref{fig:viewer_radial_menu} were presented in seven categories: \textit{I - Speed}, \textit{II - Collectivity}, \textit{III - Interpersonal Distance}, \textit{IV - Socialization and Isolation}, \textit{V - Hofstede Cultural Dimensions}, \textit{VI - Big-Five personalities} and \textit{VII - Emotions}. in that example, the OCEAN personalities of \textit{Agent6} (highlighted in red) were represented in a graphical way, considering the max value of each dimension.

\begin{figure}[!htb]
    \centering
    \subfigure[fd][Oblique view]{\includegraphics[width=0.485\linewidth]{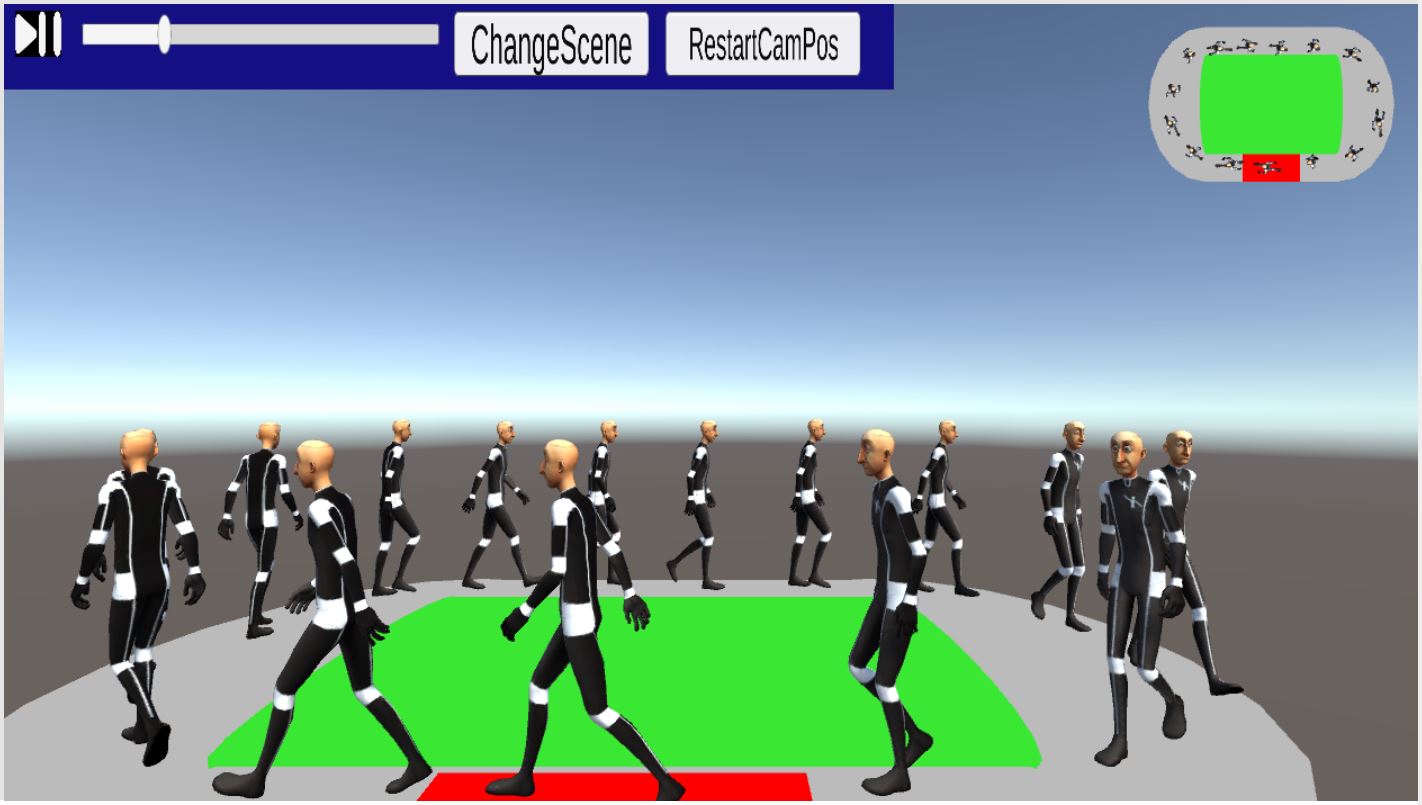}} \hfill
    \subfigure[fd][First-person view]{\includegraphics[width=0.485\linewidth]{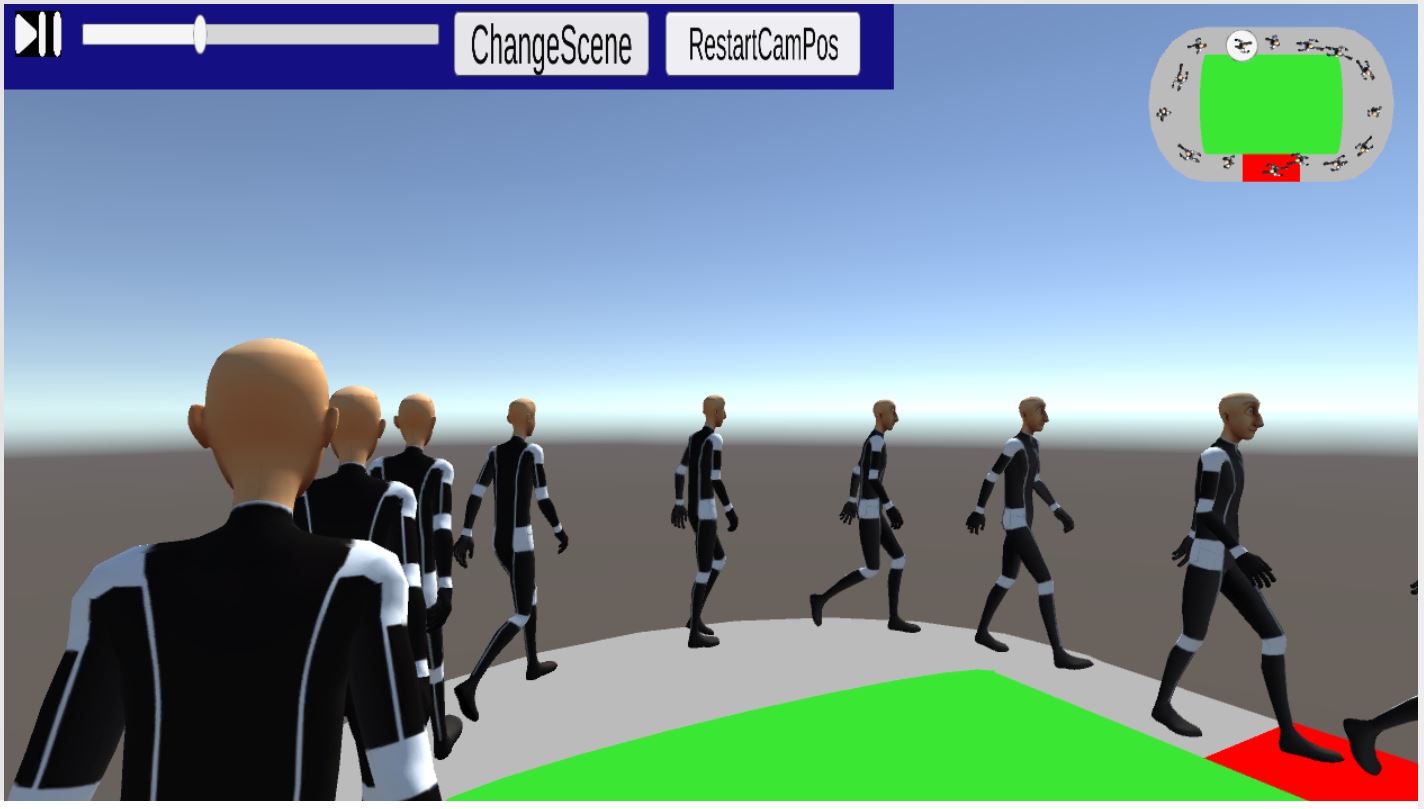}}
    \caption{\textbf{Visualization from a video of the FD experiment}: different angle of visualizations: (a) oblique view and (b) first-person view.}
    \label{fig:viewer_fd}
\end{figure}

In addition to viewing spontaneous videos, the viewer also have a module which accepts videos from controlled experiments, such as the Fundamental diagram experiment. Figure~\ref{fig:viewer_fd} shows the visualization from a video of the FD experiment, recorded in Brazil with 15 agents. In Figure~\ref{fig:viewer_fd}(a), the oblique view is shown, where the user has a more general view of the experiment. In Figure~\ref{fig:viewer_fd}(b), the first person view is shown, where the user can feel part of the experiment, with the view of one of the agents (highlighted in the top view area). In both cases, the user can see a top view of the experiment in the upper right corner of the figures.

\section{Final Considerations}
\label{sec:final_considerations}

We proposed a software called \emph{GeoMind}, which serves as a tool to detect and analyze geometric dimensions in terms of personality, emotion, cultural aspects among other features about pedestrians and groups of people in the video scene. The software was tested with Cultural Crowds dataset (available at \emph{GeoMind}'s website) and results are very satisfactory. This process of extracting information from pedestrians is very useful to better understand people's behavior, allowing analysis, comparisons and even the use of output files in other applications as games and simulations.

Finally, we present the development of a crowd-cultural data viewer, which allows the users to analyze and extract pedestrian behavior information. With three modes of visualizations (\textit{first-person}, \textit{oblique} and \textit{top view}), the viewer can be use as a complement to \emph{GeoMind} software.

As future work we intend to add new features to \emph{GeoMind}, such as new features and analysis on pedestrians and possibilities for detection and tracking of pedestrians embedded in the software, eliminating the need for external tracking files.

\section{Data Availability}

\textit{GeoMind} is available for download at \url{https://www.rmfavaretto.pro.br/geomind}. Supporting data is also available on request: please contact the authors.

\section{Conflicts of Interest}

The authors declare no potential conflicts of interest.

\section{Funding Statement}

Office of Naval Research Global - ONRG; and Conselho Nacional de Desenvolvimento Científico e Tecnológico - CNPQ [grant number 305084/2016-0].

\bibliographystyle{abbrv}

\bibliography{refs}
\end{document}